\documentclass[runningheads,a4paper]{llncs}

\usepackage{amssymb}
\usepackage{graphicx}
\usepackage{tikz} 
\usepackage[small]{caption}
\usepackage[ruled]{algorithm2e}
\usepackage{booktabs}
\usepackage{amsmath}
\usepackage{mathabx}
\usepackage{dsfont}

\usepackage{url}
\urldef{\mailsa}\path|{tuennermann, hennig, silbernagel, mertsching}@get.upb.de|

\newcommand*\circled[1]{\tikz[baseline=(char.base)]{
            \node[shape=circle,draw,inner sep=0.6pt] (char) {\small{#1}};}}

\begin{document}

\mainmatter  

\title{A Prototyping Environment for Integrated Artificial Attention Systems}
\titlerunning{Prototyping Integrated Artificial Visual Attention Systems}

\authorrunning{T{\"u}nnermann et al.} 

%
%
\author{Jan T{\"u}nnermann \and Markus Hennig \and Michael Silbernagel \and B{\"a}rbel Mertsching}
%



\institute{GET Lab, University of Paderborn, Pohlweg 47--49, 33098 Paderborn, Germany
\mailsa\\
\url{http://getwww.upb.de}
}

%
%

\maketitle

\begin{abstract}
Artificial visual attention systems aim to support technical systems in visual tasks by applying the concepts of selective attention observed in humans and other animals. Such systems are typically evaluated against ground truth obtained from human gaze-data or manually annotated test images. When applied to robotics, the systems are required to be adaptable to the target system. Here, we describe a flexible environment based on a robotic middleware layer allowing the development and testing of attention-guided vision systems. In such a framework, the systems can be tested with input from various sources, different attention algorithms at the core, and diverse subsequent tasks.   
\end{abstract}

\section{Introduction}
The robot operating system (ROS) is a popular robotics middleware. Among other features, it provides hardware abstraction, package management and a communication layer based on the publisher-subscriber paradigm allowing interaction of different components (ROS nodes) \cite{Quigley2009}. Even though the aim of ROS is simplifying system integration in robotics, it is highly useful in the development and testing of general purpose attention systems. With the concept of messages of certain types, which are published and subscribed by the components, an interface is available that allows a convenient exchange of components.

\begin{figure*}[h!t]
\centering
\hspace{-1.1cm}
\includegraphics[width=.8\textwidth]{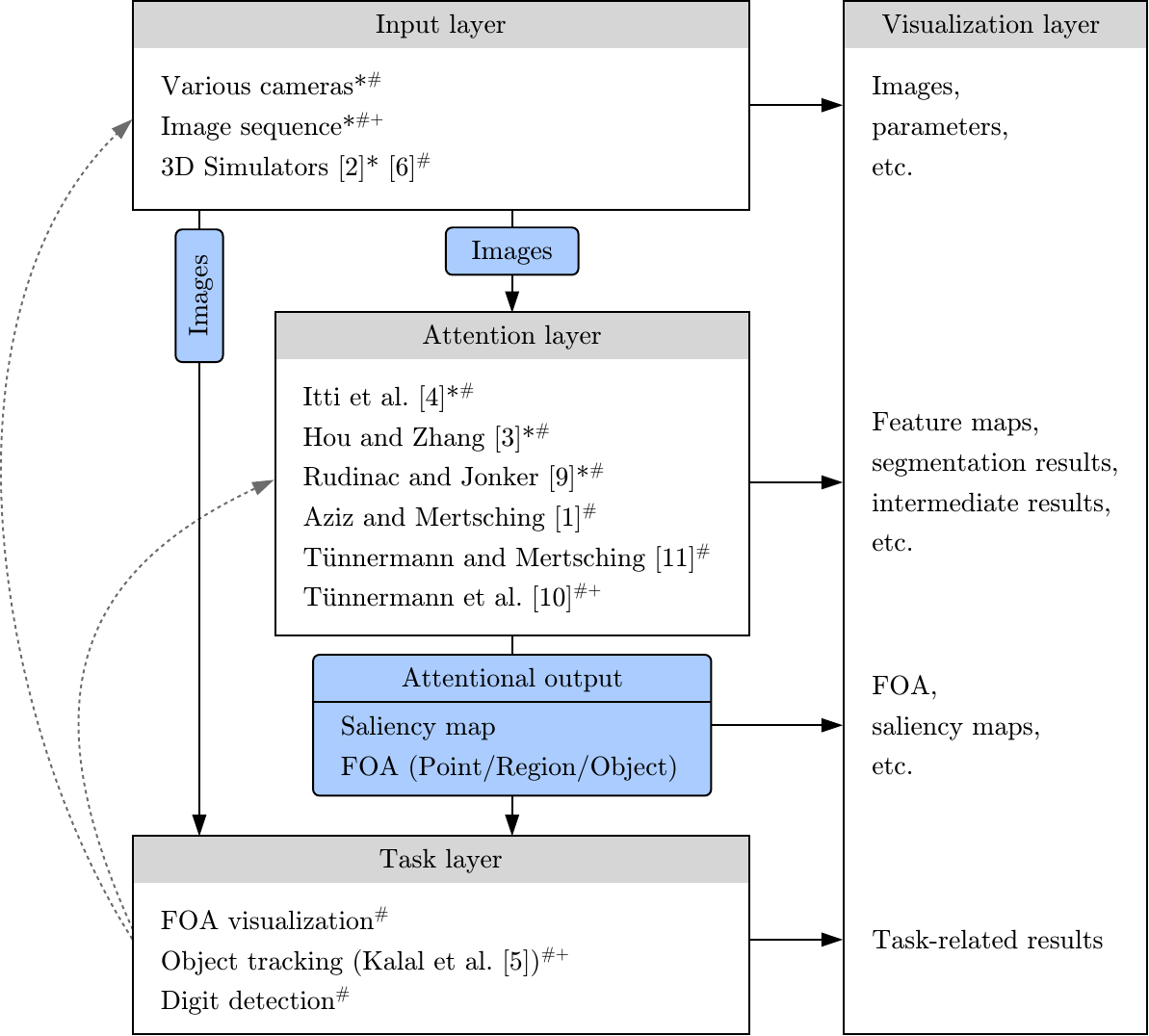}
\caption[]{Proposed ROS-based environment. Besides the different layers processing the data, the main messages, which are subject to synchronization, are shown (blocks with rounded corners; messages for the visualization are not shown). Inside the blocks representing the layers, exemplary instances of components are a listed. Publicly available components (as a part of ROS or public ROS packages) are marked with ``*''. Those marked with ``${}^{\#}$'' have been explicitly used in such an environment in our lab. ROS implementations of the models described in \cite{Itti1998}, \cite{Hou2007} and \cite{Rudinac2010} were obtained from the ROS repository \cite{Saliency_detection} and were adapted to allow for synchronization of input and results. Nodes marked with ``${}^{+}$'' are part of the example described in section \ref{example}. The dashed gray arrows represent feedback loops which are subject of future investigations.}
\label{system}
\end{figure*}

We propose an architecture consisting of an \textit{input layer} publishing images, which can be obtained from various sources. The \textit{attention layer}, which is the core of the system, performs the selective mechanisms on the images and forwards them to the \textit{task layer} (see fig. \ref{system}). Especially during the development phase, an additional \textit{visualization layer} is useful to convert the output at various stages of the system into a form which is comprehensible for the developer. In the following, we describe the layers and the contained components consisting of ROS nodes.


\nocite{Aziz2008}
\enlargethispage{1\baselineskip}

\section{The Proposed Architecture}\label{proposed}
The \textbf{input layer} performs the image acquisition. Inside the input layer block shown in fig. \ref{system}, several examples are listed. Images can be obtained from various cameras (such as webcams, pan--tilt--zoom network cameras, or cameras specialized for computer vision) or fed into the system from recorded images or videos. For many cameras, drivers exist in the form of ROS packages. Tools for feeding images or videos into the ROS system are also available. Furthermore, 3D simulators such as SIMORE \cite{Kotthauser2010} and Gazebo \cite{Gazebo} have been adapted to publish images in ROS. Additional to image acquisition, preprocessings, such as smoothing or resizing the images, are possible parts of the input layer.

The \textbf{attention layer} processes the images by applying attention algorithms (see fig. \ref{system} for a list of algorithms which have been tested as ROS nodes). Some algorithms, such as the spatiotemporal approach described in \cite{Tuennermann2013}, may require the collection several images to form a spatiotemporal context. Different algorithms provide different data at the output end: While most approaches provide saliency maps and a punctual focus of attention (FOA), others additionally contain areal information such as region outlines or bounding boxes. The latter are considered as ``Region-FOA'' and ``Object-FOA'' in fig. \ref{system}, indicating that they can cover partial objects (regions) or complete objects.

The \textbf{task layer} contains post-attentional tasks. Examples are object tracking or letter detection components, which are initialized with data from the attentional system. Depending on the task, the attentional output, supplementary processing, such as thresholding saliency maps to obtain bounding boxes, may be required.

The \textbf{visualization layer} can contain standard tools of ROS (e.g. the image viewer) or custom components (e.g. the volume viewer; see fig. \ref{screenshot} \circled{2}) that receive input from other layers.

\section{Example Instances}\label{example}
\begin{figure*}
\centering
\includegraphics[width=0.81\textwidth]{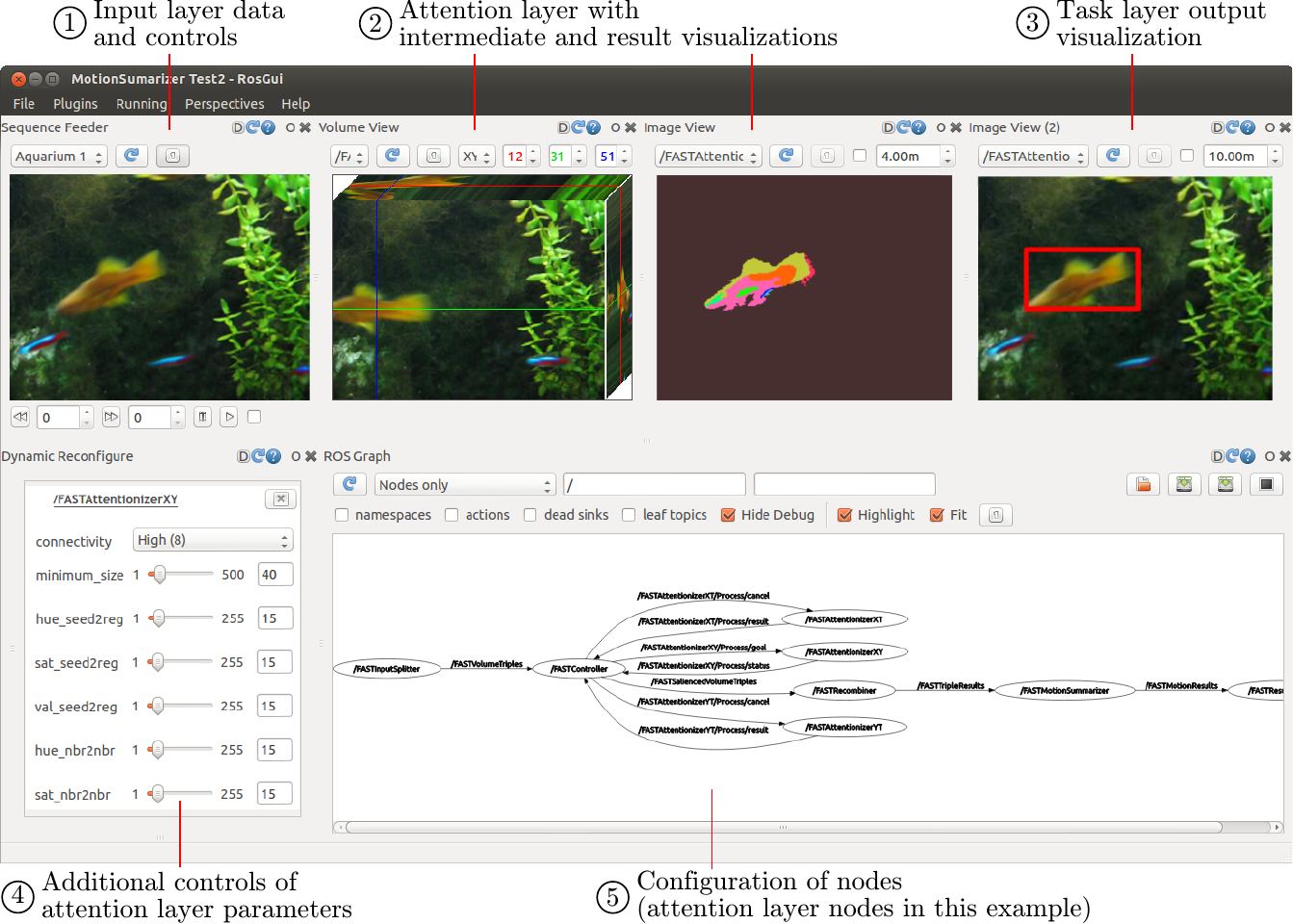}
\caption[]{ROS user interface showing a test configuration of the proposed environment. An image sequence is fed into the system \circled{1}, processed by the model reported in \cite{mainpaper} \circled{2}, and subsequently forwarded to the object tracker proposed in \cite{Kalal2012} \circled{3}. Additionally, ROS tools for parameterization  \circled{4} and system visualization \circled{5} are included.}
\label{screenshot}
\end{figure*}

Fig. \ref{screenshot} depicts an example instance of the proposed environment, which is used to test a prototypical integration of the system proposed in \cite{mainpaper} with an object tracker. This \textit{Tracking-Learning-Detection} (TLD) \cite{Kalal2012} algorithm allows real-time tracking of objects. The object of interest is defined by a bounding box in one single frame and without any previous training.

Because it is advantageous for the prototyping when results can be reproduced, a ROS node that feeds an image sequence into the system is used in the input layer. 
A network of ROS nodes, which perform different sub-task within the attention system, populates the attention layer. The output layer contains nodes to convert the object-based FOA into the bounding box and the tracker itself. The components used in this example are marked with ``${}^{+}$'' in fig. \ref{system}.

The purpose of this setup is the investigation of methods to use the attentional output to initialize the tracker. Research questions include: how to obtain a bounding box, when to start tracking, and when to stop tracking and consider a new FOA. 

\section{Outlook}\label{outlook}
Future work will be directed at integrating top-down influences. Many attention systems offer rudimentary support for these. However, feedback connections (gray arrows in fig. \ref{system}) from the task to earlier layers (including possible adjustment of preprocessings in the input layer) have not been sufficiently investigated.

Moreover, the flexible environment suggested here allows the definition of a framework for strict quantitative testing: On the one hand, the attention layer may be encapsulated by a system that provides input and compares the results to ground truth; in this way different attention algorithms can be compared. On the other hand, such an evaluation environment could also encompass an attention and a task layer, allowing to operationalize the performance of attention systems via a high-level tasks.

\bibliographystyle{splncs03}
{\scriptsize
\bibliography{bibliography}}
\enlargethispage{2\baselineskip}

\end{document}